\documentclass[12pt]{article}
\usepackage{graphicx} 
\usepackage[a4paper, margin={2.5cm} 
]{geometry}
\usepackage{amsmath, amssymb}
\usepackage{amsthm}
\usepackage{float}
\usepackage{comment}
\usepackage{placeins}
\usepackage{array}
\usepackage{booktabs}
\usepackage{hyperref}
\usepackage{linguex}
\usepackage{setspace}
\usepackage[all]{nowidow}
\usepackage[table]{xcolor}
\usepackage{txfonts}
\usepackage{pxfonts}

\newcommand{\pe}[1]{\textcolor{black}{#1}}
\newcommand{\ts}[1]{\textcolor{black}{#1}}

\title{Fairness Interventions in Classification:\\A Study on AI Explainability\thanks{Penultimate version.}}

\author{\small Thomas Souverain\thanks{CEA-Saclay \& Institut Jean-Nicod (CNRS, ENS-PSL, EHESS), Paris, France.}\and 
\small Paul \'Egr\'e\thanks{IRL Crossing, CNRS, Adelaide, Australia / ENS-PSL, Paris, France.}}
\date{}

\emergencystretch=\maxdimen
\graphicspath{ {./Images/} }

\newcommand{\fairdream}[0]{{\sc FairDream}}
\newcommand{\var}[1]{{\tt #1}}
\newcommand{\GridSearch}[0]{{\sc GridSearch}}

\begin{document}

\maketitle







\vspace{-1cm}
\begin{abstract}
\noindent
{This paper presents a philosophical and experimental study of fairness interventions in AI classification, centered on the explainability and transparency of corrective methods, and on the opposition between two fairness criteria, namely Demographic Parity and Equalized Odds. Our main argument is that even as a gap in Demographic Parity is used to diagnose inequality between groups, Equalized Odds constitutes a more reliable fairness criterion to guide bias correction in classification. To establish this, we present \fairdream, a fairness package 
intended
for lay users, whose mechanism increases the model's weights of errors on disadvantaged groups. To justify \fairdream's results, we analyze its reweighting algorithm, and we present the results of a benchmark experiment in which we compare \fairdream\ with a distinct in-processing correction method that enforces Demographic Parity more drastically, the \GridSearch\ method. We then propose a normative justification of Equalized Odds, with a discussion of the criterion's limitations. We draw on the structural similarity between \fairdream’s results and a version of Simpson’s paradox to justify conditioning on true labels in counterfactual evaluations of fairness.
}



\bigskip 
 \end{abstract}
 

{\small  \textbf{Keywords:} 
    Fairness ;
    Classification ;
    Equalized Odds ;
    Demographic Parity;
    Calibration;
    Grid Search ; Simpson Paradox; Machine Learning; {Explainable AI}

\sloppy

\newpage

\section{Introduction}



Determining whether an algorithm provides ``fair'' predictions is a challenging task, {although one that is of central practical and ethical importance for society.} The difficulties pertain in part to the fact that different notions of fairness exist and compete with each other. On the theoretical side, plethora of statistical metrics exist to quantify the extent to which the predictions of an AI model meet the user's expectation on fairness, in particular regarding a minority that must not be harmed \cite{hellman2020measuring} \cite{mehrabi2021survey} \cite{wan2023processing}. Yet, attempts to compare these algorithmic fairness metrics face impossibility theorems, implying that the various fairness criteria proposed cannot be jointly satisfied in all cases (viz. 
\cite{kleinberg2016inherent, hellman2020measuring, hedden2021statistical}). 


As an illustration of this, some concrete cases have shown us that the recommendations of an AI may be viewed as fair by its designers, but as unfair by external evaluators. The case of COMPAS (Correctional Offender Management Profiling for Alternative Sanctions), an algorithmic tool intended to score recidivism risk, is a telling example. Notoriously, the ProPublica 2016 report accused COMPAS of racial bias \cite{propublica2016compas}, by showing a higher rate of false positives in African Americans for ``high-risk'' scores and a higher rate of false negatives in Caucasians for ``low-risk'' scores. In response, the designers of COMPAS (Northpointe) replied that ProPublica artificially built these categories (by setting a boundary at 5/10 of recidivism risk), and that for each different risk level, the statistical performance of COMPAS (as measured by the \hyperlink{https://scikit-learn.org/stable/modules/generated/Scikit-learn.metrics.roc_auc_score.html}{ROC-AUC}, see section \ref{section:definitions}) was actually the same across both groups, and even slightly more favorable to black defendants \cite{dieterich2016compas}.

{But plurality of metrics is arguably not the main challenge for fairness. A more common obstacle is \textit{opacity} concerning the mechanism behind a classificatory verdict.} 
{This concerns cases in which the decision taken by an AI or by a human is issued without the user having any idea of the principles behind it. Examples could be multiplied: why did the system not give me the option to go to university when I thought I had the credentials? (see \cite{parcoursup2023senat}) Why was my loan refused when I think I could refund it? (\cite{purificato2023use})  Although no classification scheme may be immune to error, we believe that errors can at least be mitigated when the user has access to the principles behind the decision, if only because they can question them and suggest amendments.} 

In this paper, we propose to look at the problem of defining a transparent and explainable AI system for inequality detection and correction. 
In relation to that, we draw attention to two fairness criteria (\cite{mayson2018bias,castro2019s,long2021fairness,hedden2021statistical,barocas2023fairness}), which we describe more formally in the next section:

\begin{itemize}
\item \pe{\textbf{Demographic Parity}: equalize the overall positive rate across groups.}
\item \pe{\textbf{Equalized Odds}: equalize true positive rates and false positive rates across groups.}
\end{itemize}

While these are by no means the only fairness metrics worthy of consideration,\footnote{Another one is Calibration: equalize the ratio of true positives to predicted positives across groups, at identical decision thresholds. Calibration is for us less central to the normative debate concerning the redistribution of resources or true labels. To complete the present perspective, we consider the constraint in more detail in the Appendix to this paper.} our reasons for focusing on Demographic Parity and Equalized Odds are twofold. On the one hand, demands of Demographic Parity are real in various segments of society, in relation to gender, age, or ethnicity. Demographic Parity has thus been interpreted as a legal requirement, e.g. on EU Non-Discrimination Law \cite{wachter2020bias} or for the risk assessment mandated by Pennsylvania \cite{selbst2019fairness}. On the other hand, Equalized Odds corresponds to the idea of error rate parity of a classifier \cite{barocas2023fairness}, which means that the classifier treats different groups in the same way. Violations of this criterion typically appear as cases of injustice, as in the case exposed by ProPublica against COMPAS.


In this paper, we propose to cast light on the tension between these two fairness criteria by building a specific AI model for inequality detection and correction, which we call \fairdream. Thus, as with other approaches to AI fairness developed in diverse domains, including medicine \cite{chen2023algorithmic}, law \cite{lagioia2023algorithmic}, or mortgage lending \cite{lee2021algorithmic}, we anchor our philosophical analysis to a particular case study, to draw general lessons from the way in which bias can be detected and corrected using AI. More specifically, we use our case study to connect the issue of AI fairness with the issue of explainability in AI. Our motivation for this approach is the following: the main goal of the package \fairdream\ is to give lay users resources to detect and to correct for inequality. For the detection step, we use gaps in Demographic Parity as alerts for corrections the user may want to perform. For the correction step, however, \fairdream\ proceeds to narrow such gaps in accordance with the constraint of Equalized Odds, so by conditioning the objective on the true labels. The explanation for this behavior is unobvious, because it fundamentally depends on the way the loss function is set up in \fairdream, using reweighting rather than direct penalties. We first proceed to explain this behavior, in particular by comparing the correction mechanism used by \fairdream\ to closely related mechanisms that implement the constraint of Demographic Parity more drastically  using a different loss function. Then we justify it with normative arguments in favor of Equalized Odds as a fairness criterion. In particular, we draw on the close connection between the results of \fairdream\ and a version of Simpson's paradox to justify conditioning on true labels in counterfactual evaluations of fairness.


The way the paper proceeds is as follows: 
Section \ref{section:DP_EO} starts with some general definitions used throughout the paper, and introduces the opposition between Demographic Parity and Equalized Odds. Section \ref{section:package} then introduces the package \fairdream\ and in particular the various parameters that the user can set and manipulate in order to detect and correct for inequalities. It shows on a specific dataset and example the way in which \fairdream's correction enforces the property of Equalized Odds, even as the fairness objective set by the user is to reduce gaps in Demographic Parity. Section \ref{section:explain_results} then explains \fairdream's mechanism, and Section \ref{section:grid_search_benchmark} presents the results of a systematic benchmark experiment with the \GridSearch\ algorithm of \cite{agarwal2018reductions} that remains state of the art in this domain. In section \ref{section:decision_rule} we discuss the limits of conditioning fairness interventions on true labels, and in section \ref{sec:simpson} we take advantage of an analogy between \fairdream's results and the Simpson paradox to justify conditioning on true labels. {Section \ref{sec:conclusion} concludes with broader lessons regarding the way in which the explanatory steps taken here make the model trustworthy.} Finally, in an Appendix we discuss the notion of Calibration and show how \fairdream\ performs relative to that metric.

   \section{Demographic Parity vs Equalized Odds}\label{section:DP_EO} 

   In this section, we present the two central fairness metrics discussed in this paper, namely Demographic Parity and Equalized Odds. First, we introduce some basic concepts and notations that will be used throughout the paper and that are needed to state these definitions. 

    \subsection{Basic Concepts}\label{section:definitions}

    In a classification problem, the algorithmic model makes a \textbf{prediction} $\hat{Y}$ about a \textbf{variable} $Y$ of interest. Each value of $Y$ corresponds to a different class. 
    When the classification problem is binary as in this paper (such as predicting whether a person earns more than \$50,000 or not), $Y$ and $\hat{Y}$ take the Boolean values $0$ (for earning less than \$50,000) or $1$ (for earning more). 
    Similarly, we write ``$A=1$'' and ``$B=1$'' to indicate whether a person is a member of group $A$ or of group $B$. 

    \begin{table}[h]
        \centering
        \begin{tabular}{c|c|c}
         &  $\hat{Y}=1$   & $\hat{Y}=0$  \\
         \hline
      $Y=1$   & 
      $TP$ & 
      $FN$ \\
      \hline
      $Y=0$  &  
      $FP$   
      & $TN$ \\ 
        \end{tabular}
        \caption{True Positives, False Negatives, False Positives, True Negatives}
        \label{tab:cm}
    \end{table}
    
    Different metrics exist to quantify the algorithm's classification performance, which can be defined in terms of the categories represented in Table \ref{tab:cm} of True Positives (TP), False Negatives (FN), False Positives (FP), and True Negatives (TN). The following definitions are all relative to some group $G$, reference to which is left implicit after the first definition for brevity. We use the conditional probability notation $p(E|F)$ to denote the proportion of $F$ that satisfies $E$.
    
    
        
    The \textbf{Overall Positive Rate} (OPR) of a prediction $\hat{Y}$ in a group $G$ is defined as the proportion of individuals predicted to have $Y$, i.e. $p(\hat{Y}=1|G=1)$, it is equal to $\dfrac{{TP+FP}}{{TP+FN+FP+TN}}$. 

        The \textbf{True Positive Rate} (TPR), also called \textbf{Recall}, or \textbf{Sensitivity}, is $p(\hat{Y}=1|Y=1)$. 
        It is the proportion of true positives that are predicted to be positive, that is $\dfrac{TP}{TP+FN}$. 

        
           
    The \textbf{True Negative Rate} (TNR), also called \textbf{Selectivity}, is defined as $p(\hat{Y}=0|Y=0)$, it is equal to $\dfrac{TN}{TN+FP}$. 

    The \textbf{False Positive Rate} (FPR) is defined as $p(\hat{Y}=1|Y=0)$, it is equal to  $\dfrac{FP}{FP+TN}$, or equivalently to 1-TNR. 
    
        The \textbf{False Negative Rate} (FNR) is defined as $p(\hat{Y}=0|Y=1)$, it is equal to $\dfrac{FN}{FN+TP}$ or equivalently  to 1-TPR.

    The \textbf{Precision}, also called \textbf{Positive Predictive Value}, is the proportion of predicted positives that are true positives, $p({Y}=1|\hat{Y}=1$), that is $\dfrac{TP}{TP+FP}$. 

        The \textbf{Proportion Correct} is the proportion of true positives and true negatives compared to all predictions, that is $\dfrac{TP+TN}{TP+FN+FP+TN}$.

        The Proportion Correct is sometimes called the Accuracy of a model, although we will use the term ``accuracy'' more broadly to refer to how well a model performs relative to other performance metrics.
Three other common performance metrics will be used in what follows: 

The \textbf{ROC-AUC} (Receiver Operating Characteristic Area Under the Curve) is the area under the curve plotting, for every possible classification threshold (viz. the age value chosen as a predictor variable for income; or a probabilistic value if the model outputs probabilities), the relation between its corresponding True Positive Rate (y-axis) and False Positive Rate (x-axis).

The \textbf{PR-AUC} (Precision-Recall Area under the Curve) is the area under the curve plotting the relation between Precision (y-axis) and Recall (x-axis).\medskip
Precision and Recall calculate the proportion of True Positives relative to different reference classes, and so a standard way of aggregating them for assessing performance is to compute their harmonic mean:\footnote{The harmonic mean between two numbers $a\geq 0$ and $b\geq 0$ is $0$ when both $a$ and $b$ equal $0$, and is equal to $\dfrac{2ab}{a+b}$ otherwise.} \medskip

    The \textbf{F1-score} is the harmonic mean between Precision and Recall, it is equal to $\dfrac{TP+TP}{TP+FN+FP+TP}$. 
    
The F1-score may thus be seen as a modified version of the Proportion Correct in which only True Positives matter and not True Negatives.

    


\subsection{Two Fairness Metrics}

    The issue we are dealing with in this paper is how to correct biases in classifiers and so how to produce outcomes that ensure fairness. 
    For example, assuming two different groups $A$ (younger clients, females, etc) and $B$ (older clients, males, etc), and that the algorithm makes a prediction regarding some variable of interest (such as whether the individual earns above a certain yearly income, say \$50,000), how can we ensure that the algorithm treats $A$ members and $B$ members in the same way?

We consider two main criteria for fairness in binary classification. First, an algorithmic classification method may be considered fair if it predicts the same percentage of individuals earning above \$50,000 across the two groups. That corresponds to an objective of {Demographic Parity} \cite{dwork2012fairness,wachter2020bias}.

    An algorithm satisfies \textbf{Demographic Parity} between two groups $A$ and $B$ with respect to $Y$ provided:
    $$p(\hat{Y}=1 |A=1)=p(\hat{Y}=1|B=1)$$
    
Demographic parity is not sensitive to the base rates, that is to the proportion of individuals who actually earn more than \$50,000 in each group. All that matters to fairness on that criterion is the \textit{terminus ad quem}, namely to achieve the same Overall Positive Rate (OPR) across both groups.

A distinct and competing criterion corresponds to the idea that errors should be distributed in the same way across the two groups when considering the true labels. This corresponds to a metric called {Equalized Odds} \cite{hardt2016equality, barocas2023fairness}, which can be presented as follows:

An algorithm satisfies \textbf{Equalized Odds} between two groups $A$ and $B$ with respect to $Y$ provided:
$$\left\{
    \begin{array}{ll}
        p(\hat{Y}=1|Y=1\wedge A=1)=p(\hat{Y}=1| Y=1 \wedge B=1)\ \\
        p(\hat{Y}=1|Y=0\wedge A=1)=p(\hat{Y}=1|Y=0 \wedge B=1)
    \end{array}
\right.$$

What Equalized Odds says is that the True Positive Rate (TPR) and False Positive Rate (FPR) should be the same in each group. 
According to this criterion, the \textit{terminus a quo} matters too, meaning that the base rates ought to be taken into consideration for prediction.


In the rest of this paper, we defend the idea that consideration of the true labels is a safeguard for fairness, and that all things considered Equalized Odds is a more adequate fairness criterion than Demographic Parity. To establish this, in the next section we deploy an algorithmic method of fairness intervention, the package \fairdream, and we test it on actual data. 



\section{The package {\sc \fairdream}}\label{section:package}


The main purpose of the package \fairdream, \pe{originally created in response to an AI fairness contest run by the Monetary Authority of Singapore (MAS)},\footnote{For more details, see the hackathon launched by the MAS under the banner of \href{https://www.mas.gov.sg/schemes-and-initiatives/veritas}{Veritas Initiative}.} is to give lay users, without any background in data science or in theories of fairness, an insight into the baseline AI model, and subsequently, a handle on the process of correction. Here, we first present the general procedure followed by \fairdream. We then introduce the dataset on which \fairdream\ was tested, and show that the results obtained are fundamentally constrained by Equalized Odds.



\subsection{General Principles}


\fairdream\ takes as input an already trained predictor (some baseline machine-learning model, in our case an XGBoost model) over a given dataset, and it asks the user to select a ``fairness objective to compare groups'' (see Figure \ref{fig:fairdream_menu}). The ``fairness objective'' refers to the metric used to detect a gap or disparity among groups, which the user seeks to minimize at the correction step.

The procedure implemented in \fairdream\ involves two steps, a detection step and a correction step. At the detection step, \fairdream\ outputs an alert once it has detected a disparity. At the correction step, \fairdream\ produces several alternative models and selects the one that minimizes the disparity detected.

\begin{figure}
    \centering
    \includegraphics[width=0.6\linewidth]{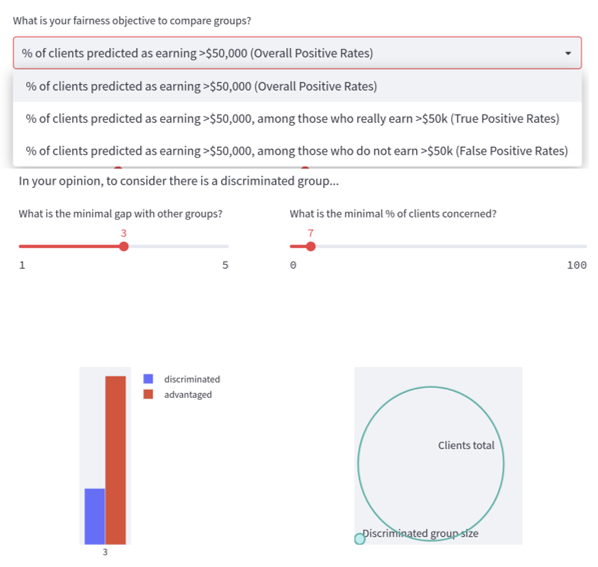} 
    \caption{Menu options in \fairdream: fairness objective (top) and disparity factor (bottom)}
    \label{fig:fairdream_menu}
\end{figure}

\textbf{Detection} - The ``Discrimination Alerts" algorithm detects disparities in how groups are treated. Each feature is inspected. If age intervals, certain jobs, or nationalities are under-selected by the model, a discrimination alert is issued. The goal is to make the user aware of disadvantaged groups, potentially distinct from what the user believes those groups might be.


\textbf{Correction} – With these alerts, the user can form a justified opinion about imbalance between populations; and they can decide which gaps between groups of features to correct for. In our experiment, we simulated the approach of a decision-maker whose normative preference is to reduce disparities between older and younger clients.\medskip 


The way the detection process works in \fairdream\ is relative to a threshold (gap ratio) set by the user. An alert is issued if the difference in proportion of individuals predicted to have feature ${Y}$ exceeds that threshold relative to the variable of interest. For the correction step, new models are trained to maximize a tradeoff score (\var{tradeoff\_score}) between statistical accuracy (\var{stat\_score}) and fairness (\var{fair\_penalty}). Those three scores are defined as follows:

\begin{itemize}
    \item the \var{stat\_score} is the performance metric chosen by the user (Proportion Correct; ROC-AUC; PR-AUC).
    
    \item The \var{fair\_penalty} is the difference between the highest and the lowest \var{fair\_score} in the metric chosen, where the \var{fair\_score} is the metric used to evaluate the disparity between groups, which the user aims to minimize. This \var{fair\_score} metric can be the overall positive rate (OPR), true positive rate (TPR), or false positive rate (FPR). The \var{fair\_penalty} is inspired by MinMax justice approaches (\cite{rawls1999justice,barsotti2024minmax}), which seek to minimize the gap between the worst and the best treated groups relative to a given \var{fair\_score} metric.\footnote{As formalized in Section \ref{subsection:reweighting}, among $m$ groups, e.g. of age, the fair penalty gets the maximal gap $max(\var{gap\_fair\_score}(S_k))_{ k\in[1,m]}$.}
    
    \item  {The $\var{trade\_off\_score}$ is an empirical  metric that we define as the convex combination of the previous two scores: $\var{trade\_off\_score}= \alpha\times\var{stat\_score} + (1-\alpha)\times\var{fair\_penalty}$.} An $\alpha$ smaller than $.5$ indicates a preference for fairness over accuracy. This parameter too is set by the user. 
    
    \end{itemize}

In total, the choice of the gap ratio needed to trigger an alert, the statistical performance metric, the fairness objective, and finally the weight bestowed on either of the latter, are all made intelligible and left at the user's discretion in \fairdream.  



\subsection{The Census dataset}

For our experiments, we manipulated the well-known Census dataset, which aggregates data from 48,842 US citizens in 1994.\footnote{\url{https://www.kaggle.com/datasets/uciml/adult-census-income}.} The dataset includes 14 characteristics such as age, work class, years of education, marital status, and geographical origins. A fifteenth column, finally, displays a binary encoding of whether the agent earns less or more than \$50,000 a year (Cf. Figure \ref{fig:censusdataset}).

\begin{figure}[h]
\caption{A sample of the Census dataset}\label{fig:censusdataset}
\centering
\includegraphics[scale=0.31]{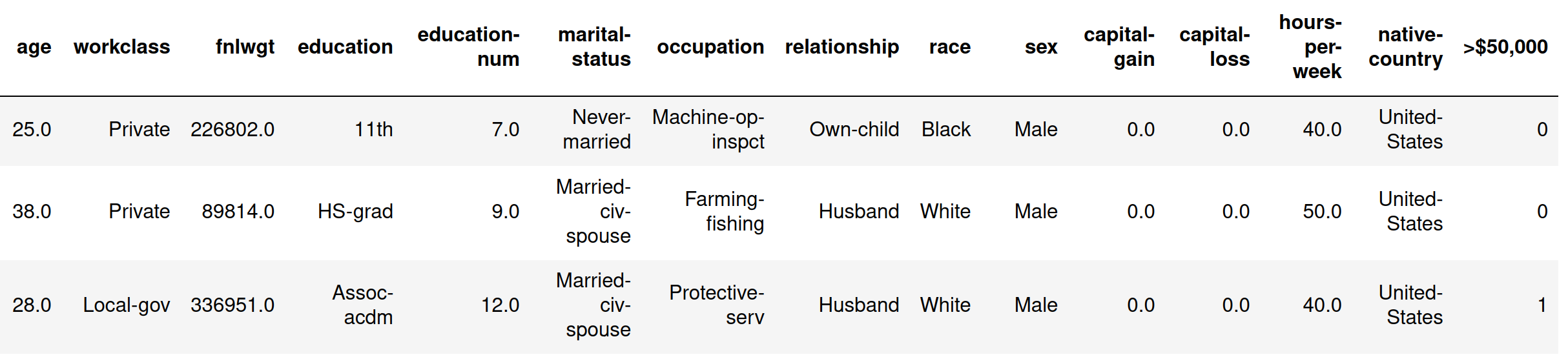}
\end{figure}

\begin{figure}[h]
\caption{Proportion of individuals earning over \$50,000 by age. The red and green lines delineate the groups 17-29 and 29-37 between which \fairdream\ detects a disparity.}\label{fig:census_wealth_by_age}
\centering
\includegraphics[scale=0.5]{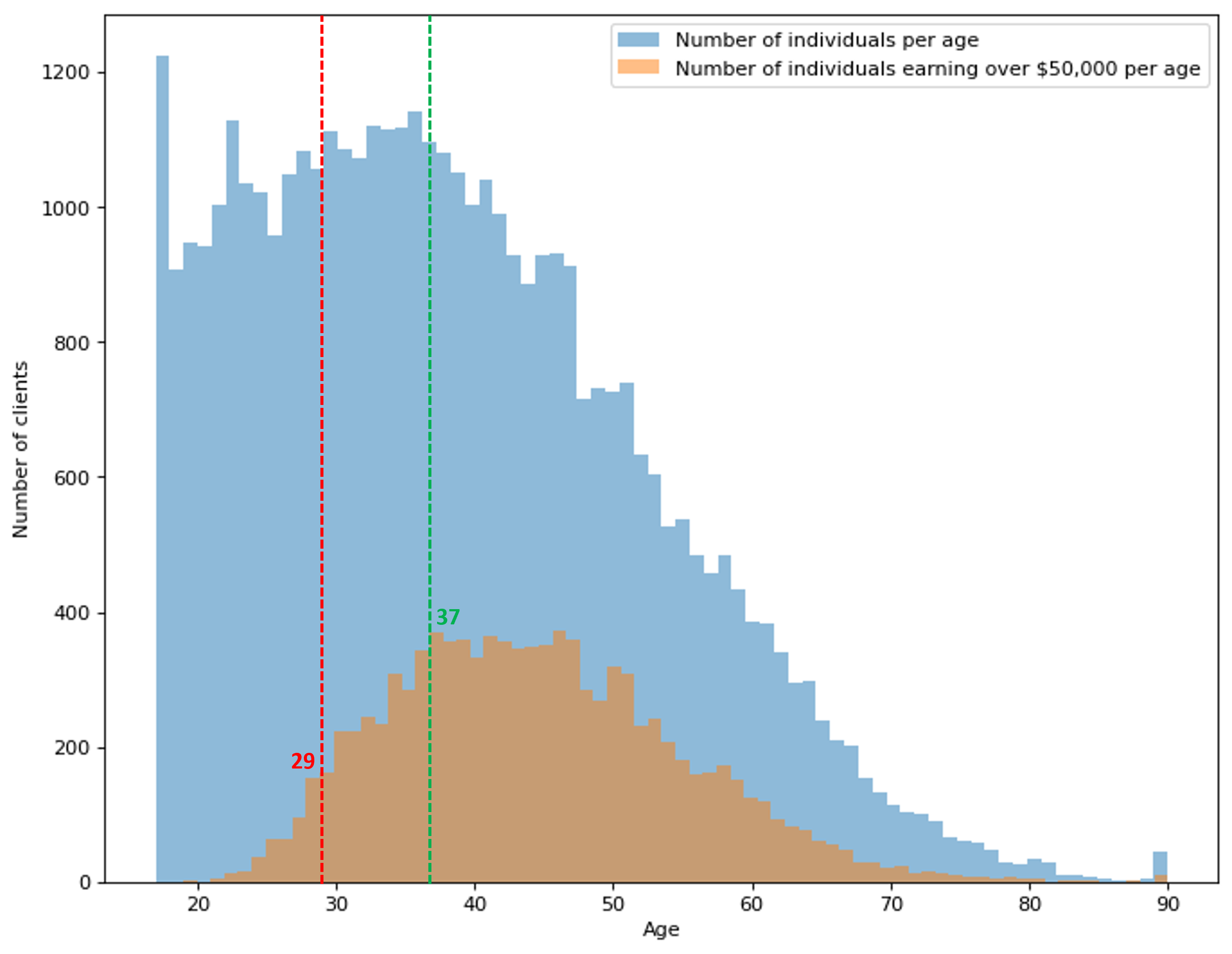}
\end{figure}


In the case at hand, the distribution of ages appears as a good predictor for the income variable (Cf. Figure \ref{fig:census_wealth_by_age}). The proportion of individuals earning over \$50,000 in the group under 29 years of age (brown area/blue area) is very small (453/11,295 individuals, i.e. 4\%). However, this proportion grows quite significantly between 29 and 37 years of age (2393/9914 individuals, i.e. 24\%). Therefore, an AI classifier might use age (among other features) as a critical factor in order to predict the wealth of individuals -- in particular, to generalize its estimates of income to data with unknown labels.

\subsection{Baseline model used}

Using the Census dataset, we first built a machine-learning model whose purpose is to minimize the error between its predictions and the actual income of people. Because the practitioner can choose different learners and different metrics of statistical performance, we built a tree-based AI model with the \href{https://xgboost.readthedocs.io/en/stable/}{XGBoost library} and evaluated it using the standard \hyperlink{https://scikit-learn.org/stable/modules/generated/Scikit-learn.metrics.roc_auc_score.html}{ROC-AUC} measure.

An XGBoost classifier --- for eXtreme Gradient Boosted trees --- is an ensemble of trees (\cite{chen2016xgboost}). Each tree tries to find the best features (splitting populations on their values) to distinguish between individuals earning more or less than \$50,000. Boosting relies on a sequence of weak trees, progressively correcting each other to reach the best predictor. In XGBoost, the computation is made faster through gradient approximation of the trees' errors, and techniques such as tree-pruning are used to find new trees faster (see \cite{shwartz2022tabular} for an overview). For stability of the following experiments, we reduced the architecture of the gradient boosted trees to a minimal tuning. By default, we set the model to 1,000 estimators, the maximal depth of each tree being 3 splits on features.

\subsection{Experiment and Results}

To set \fairdream\ in motion, we used the following parameters in our experiment {on models of the XGBoost type}. First, we used ``age'' as the variable of interest, and we set a gap ratio equal to 5. We used ROC-AUC as the statistical performance metric. {As for fairness, we set the objective to be to reduce the gap in overall positive rate (OPR) between the best treated group and the worst treated group.
}
For the tradeoff-score parameter, we set $\alpha=\frac{1}{3}$, to indicate a preference for fairness over accuracy.

Table \ref{fig:baseline} shows the results of the baseline model at the first step of the procedure. The most disadvantaged group in terms of OPR is the 17-29-year-olds (12\% selected), and the most advantaged group is the 29-37-year-olds (66\% selected).
Since the ratio of the latter to the former exceeds 5, \fairdream\ triggers an alert. At the correction step, five new XGBoost models were then trained to balance statistical performance (ROC-AUC) with the $\var{fair\_penalty}$ (in the baseline model, this penalty score is equal to $0.66-0.12=0.54$). Figure \ref{fig:fairdream_correction_selection_plot} maps the baseline model (in blue) with four of these new models, including the best model (in red) for the chosen $\alpha$. Table \ref{fig:best} shows the predictions of this model. 


\begin{table}[h]
\caption{Overall Positive, True Positive, and False Positive rates in the baseline model}
\label{fig:baseline}
\centering
\includegraphics[scale=0.80]{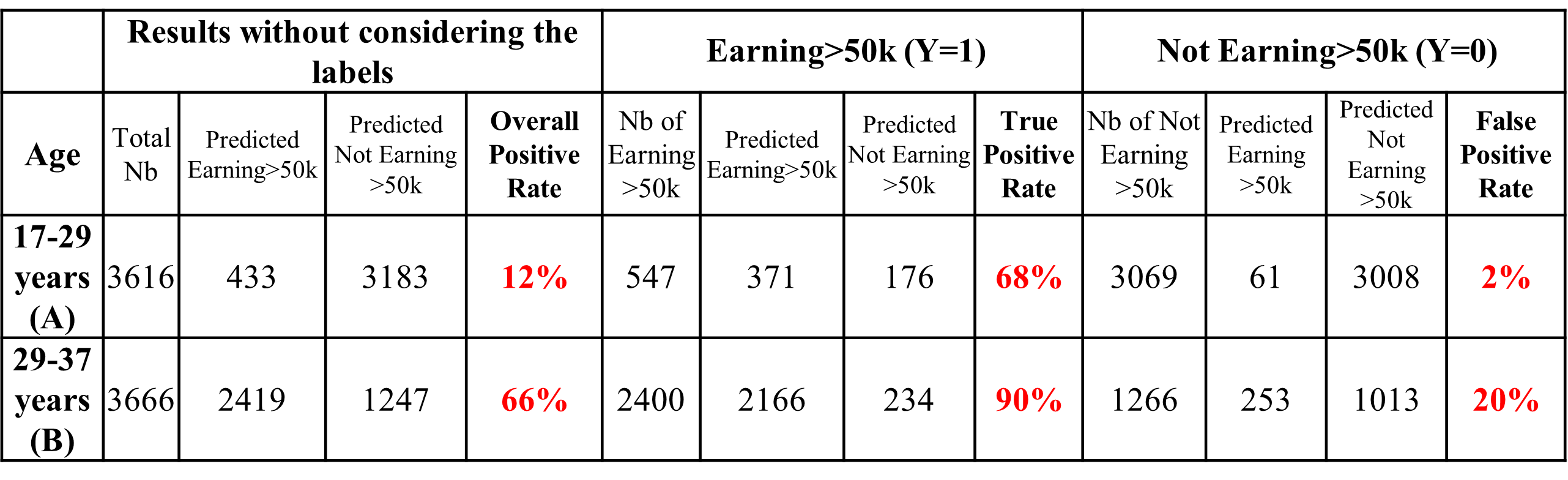}
\end{table}

\begin{figure}[t]
\caption{Selection of the best corrected model (red), based on a combined measure of statistical performance and fairness.
}
\label{fig:fairdream_correction_selection_plot}
\centering
\includegraphics[scale=0.7]{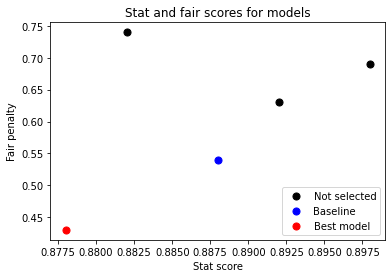}
\end{figure}

\begin{table}[h]
\caption{Overall Positive, True Positive, and False Positive rates in the \fairdream\ model}\label{fig:best}
\centering
\includegraphics[scale=0.8]{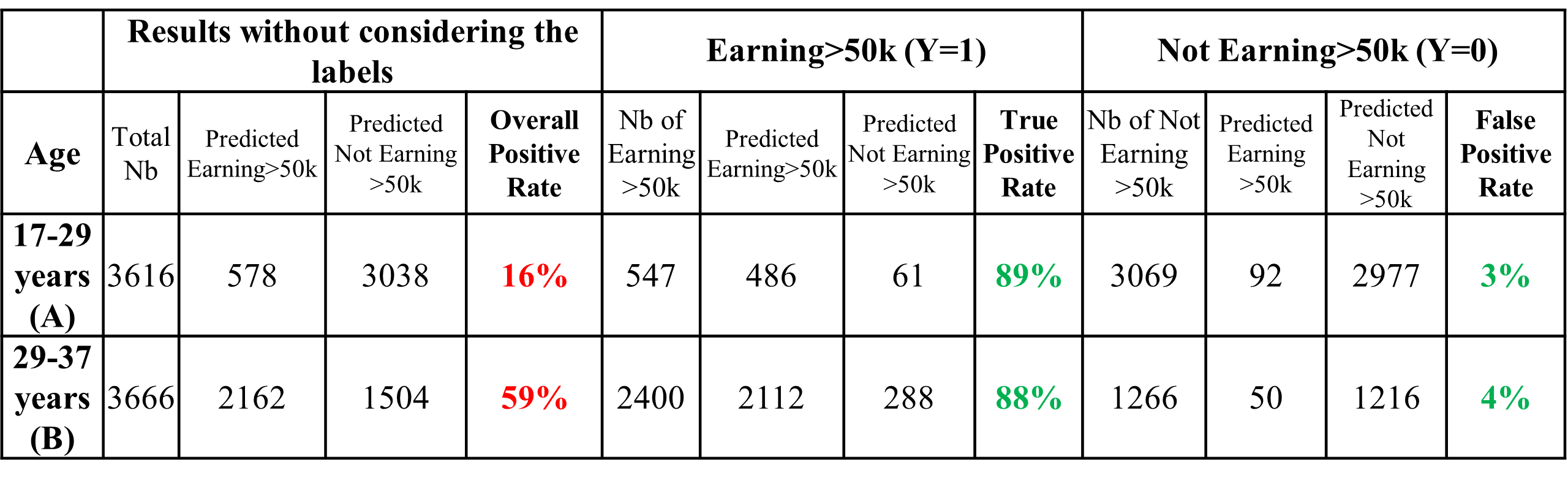}
\end{table}

At the global level, comparing Table \ref{fig:best} and Table \ref{fig:baseline} shows that the disparity between age groups in the corrected model is indeed less than 1 to 5, but the correction seems surprisingly small (16 \% vs 59 \% instead of 12 \% vs 66 \%), leaving a gap ratio of 3.7. However, when comparing by true labels we do see an important correction, since true positive rates and false positive rates come out nearly equal (88\% to 89\%, 3\% to 4\%), unlike in the baseline model (68\% to 90\%, and 2\% to 20\%). This can be interpreted as follows: while the fairness objective is to narrow the gap in OPR between the most advantaged and the most disadvantaged group, \fairdream\ does not achieve Demographic Parity, instead it comes closer to achieving Equalized Odds. In the next section, we proceed to explain this result. This explanation is centrally needed, since a lay user who only has access to the menu options represented in Figure \ref{fig:fairdream_menu} could easily misinterpret the kind of correction operated by \fairdream.

\section{Explaining \fairdream's results}\label{section:explain_results}

To explain the correction method of \fairdream, we need to say more about the landscape of “algorithmic fairness”, that is about extant methods to bring a model closer to statistical fairness criteria \cite{mehrabi2021survey}. {We describe the in-processing character of the method, and we explain the distinctive way in which reweighting works in it.} For more details, the reader can refer to the full code available in the anonymized GitHub repository: \url{https://anonymous.4open.science/r/fair-reweighting-study/README.md}.

\subsection{An in-processing method}\label{subsection:in_processing}

Once we set \pe{the goal of equalizing the percentage of clients able to repay their loan across groups,}
the fairness techniques can equalize them {\it before}, {\it during}, or {\it after} training of the model \cite{alves2021reducing}. Although they are implemented in diverse ways, these techniques mainly rely on the following architectural principles:

\textbf{Pre-processing}: Pre-processing methods intervene on the training data, to feed the future model with rebalanced or more favorable data for harmed individuals. The methods can change the labels (e.g. FairBatch \cite{roh2020fairbatch}), or the frequency of minorities in the training dataset (e.g. reweighting \cite{calders2009building}). 

\textbf{Post-processing}: 
Post-processing occurs after the model is trained. In classification, it occurs when we convert the probabilities into binary values. Some examples include adjusting the thresholds, to allow disadvantaged individuals with lowest scores to be labeled as earning \$50,000 (\cite{hardt2016equality}).
	
\textbf{In-processing}: In-processing methods generally consist in minimizing the loss function, subject to a fairness constraint, to satisfy a trade-off between the performances of the statistical and fairness metrics of the user \cite{wan2023processing}. 

\fairdream\ is an in-processing method, in which the correction step takes place during model training. Underlying this choice, we considered that intervening outside the model unnecessarily increases the opacity of an AI system. In particular, the other two methods can lead the model to learn from unrealistic samples (in the case of pre-processing), or to generate new thresholds for the same group each time the model is deployed in a new context (case of post-processing).

\subsection{ 
Reweighting for transparent processing\label{subsection:reweighting}}

In our attempt to make the method transparent to lay users, our method uses the principle of reweighting (\cite{calders2009building}. Reweighting of harmed individuals generally matches the intuition of lay users regarding compensatory justice \cite{velasquez1990justice}. That is, if the weight of an error made on a 28-year-old is three times the weight of an error on older clients, the model has to fit its parameters 
by ``paying three times more attention'' to previously harmed individuals. 
\ts{However, in \cite{calders2009building}, the weights of disadvantaged individuals were implemented by duplicating these individuals in the dataset, which was not the most efficient approach. 
Our reweighting method instead introduces the weights directly during training, treating them as errors that the classifier learns to avoid with higher priority.}

To make the correction of the model fit the lay intuition, \fairdream\ reweights groups in an ascending way. The weight of an error on a group $S_k$ grows with its previous disadvantage. The new sample weight $w_n(S_k)$ of \fairdream's model $\#n$ increases with the number of harmed individuals (linearly), and with the difference of fair scores in the baseline model (exponentially). 
For example, in \fairdream's model $\#2$, which was selected as the best one, the weight of an error made by the classifier during training on 17-29-year olds ($w_2(S_1)$=0.8) became eight times the error on 29-37-year olds ($w_2(S_2)$=0.1). 

The new  model $\#n$, $\phi_{\theta,n}$, has to learn the parameters $\theta$ that minimize the loss, and thereby maximize the statistical performance. With each new classifier $\phi_{\theta,n}$, \fairdream\ tests a new combination of weights $W_n=(w_n(S_1),\dots, w_n(S_p))$ on the $p$ groups, simply adding the multiplying weight $W_n$ inside the training loss function $L$:\footnote{$ l:\{0,1\}^2\rightarrow\mathbb{R}$ is a statistical measurement of the classification error between the target $Y_i$ of the $m$ individuals and the prediction
$\hat{Y_i}=\phi_{\theta,n}(X_i)$; $X_i$ 
is the vector of inputs; and 
$\Omega$ 
is a regularization term to avoid overfitting.}

    \begin{equation}
  \notag
  L(\phi_{\theta,n} ,\ W_n)=\ \sum_{1\leq k\leq p}\ \sum_{i\in S_k}{w_n(S_k)\times l(Y_i,\hat{Y_i})+\Omega(\theta)}
  \label{eqn:fairdream}\tag{\fairdream\ loss}
\end{equation}


Thus, instead of “leveling down” the well-off group to make its overall positive rate closer to the worse-off group, \fairdream\ focuses on enhancing the overall positive rate of the worse-off group, as done in \cite{mittelstadt2023unfairness}. For FairDream's model \#$n$, the new weight of error for each group ${S_k}$ is calculated as follows:

$w_n(S_k) = \var{rate_\_indivs_\_disadvantaged}(S_k) \times exp(n \times \var{gap_\_fair_\_scores}(S_k))$

The exponential part of the equation focuses on how far from the maximum the current fairness score (e.g., overall positive rate) of the group $S_k$ is. It resets the weight according to the disadvantage of $S_k$:

$\var{gap\_fair\_scores}(S_k) = |\var{fair\_score}(S_k) - max(\var{fair\_score}(S_1), \ldots, \var{fair\_score}(S_m))|$

The new weight $w_n(S_k)$ takes into account the difference across fairness scores, but also the number of individuals previously impacted. $\var{rate\_indivs\_disadvantaged}$ is a coefficient that stresses the number of people disadvantaged within the group, relative to the overall population. This coefficient increases as the people in $S_k$ represent a higher share of the population: 

$\var{rate\_indivs\_disadvantaged}(S_k) = \var{gap\_fair\_scores}(S_k) \times \frac{|S_k|}{\sum_{i=1}^{m} |{S_i}|}$


In summary, \fairdream\ makes the cost of errors (false negative rate and false positive rate) proportional to the structural disadvantage experienced by a group in society (e.g. women who need to be reevaluated on credit allocating, taking into account the disadvantages they face in education, recruiting, income, etc.). 


\section{Experimental comparison with \GridSearch}\label{section:grid_search_benchmark}

To investigate the proper effects of \fairdream's  correction, we compared it in a benchmark experiment with a closely related fairness method: The {\GridSearch} method of \cite{agarwal2018reductions}, also an in-processing method of cost-sensitive classification.\footnote{\url{https://fairlearn.org/main/api_reference/generated/fairlearn.reductions.GridSearch.html\#fairlearn.reductions.GridSearch}. Grid search is not specific to fairness classification: it refers to the technique of exhaustively searching the best hyperparameters of a model using a grid of pre-specified values. However, here under \GridSearch, we refer to the specific grid search method used by \cite{agarwal2018reductions} and referred to in the Fairlearn project under that name, see \cite{weerts2023fairlearn}.} The experiment's goal is to show that the convergence to Equalized Odds observed in our previous experiment is found also with other models and using other fairness objectives. Then, the goal is to understand its mechanism by contrast to a method that enforces Demographic Parity in a more drastic way. Detailed and complete results for our experiment are accessible in the benchmark repository: \url{https://anonymous.4open.science/r/weights_distortion_impact-15/}.

\subsection{The \GridSearch\ method}

In \GridSearch, statistical performance and fairness are handled separately via a constrained optimization problem. More precisely, the goal is to minimize the training loss subject to a fairness constraint $\mathcal{F}$, such as Demographic Parity, penalizing the model when the overall positive rates are not equalized across groups (within $\eta>0$):
$$\min_{\theta} L(\phi_{\theta,n} )\ \text{such that}\ \mathcal{F}(\phi_{\theta,n} )\leq \eta$$
In practice $\mathcal{F}(\phi_{\theta,n} )- \eta$ represents a vector of potential constraint violations (one per group), and the constrained problem is solved via its Lagrangian, in which $\boldsymbol{\lambda}$ is a vector of nonnegative Lagrangian multipliers. The classifier $\phi_{\theta,n}$ needs to find the parameters $\theta$ that minimize the maximum over $\boldsymbol{\lambda}$ of $L\left(\phi_{\theta,n} ,\ \boldsymbol{\lambda}\right)$:
    \begin{equation}
    \notag
    L\left(\phi_{\theta,n} ,\ \boldsymbol{\lambda}\right)=\ \sum_{i=1}^{m}\ l\left(Y_i,\hat{Y_i}\right)+\Omega(\theta)+\boldsymbol{\lambda}^\top\left(\mathcal{F}\left(\phi_{\theta,n} \right)-\eta\right)
  \label{eqn:grid_search}\tag{\GridSearch\ loss}
\end{equation}

Whereas \fairdream\ directly integrates the fairness objective through the sample weights $W_n$, \GridSearch\ uses the Lagrange multipliers to penalize violations of the fairness constraint.\footnote{For more details on the \GridSearch\ algorithm, see \cite{agarwal2018reductions}.}
\GridSearch\ starts with a grid of vectors $\boldsymbol{\lambda}$ as \fairdream\  with the weights $W_n$, which convey in an accessible way the idea of a classification sensitive to protected individuals. 

\subsection{Benchmark experiment}

We conducted the experiment with several types of models. To grant stability of the experiments, we selected the event predicted by the model (earning over \$50,000 or not) for the threshold maximizing the F1-score, commonly used in machine-learning for imbalanced classification as in Census.\footnote{\label{fn:f1}
We used the \hyperlink{https://scikit-learn.org/stable/modules/generated/sklearn.metrics.f1_score.html}{Scikit-learn metric of F1-score}.}

\begin{itemize}
    \item Gradient boosted trees (using the \href{https://xgboost.readthedocs.io/en/stable/}{XGBoost library}: 1000 estimators, with the maximal depth of each tree being 3 splits on features) -- to keep on investigating the initial type of model we used in Section \ref{section:package}.
    \item Random forest trees (using  the \href{https://scikit-learn.org/stable/modules/generated/Scikit-learn.ensemble.RandomForestClassifier.html}{Scikit-learn library}: 100 estimators, with the maximal depth of each tree being 3 splits on features) -- to introduce a lighter tree-based model.
    \item Neural networks (using the \href{https://pytorch.org/tutorials/beginner/blitz/cifar10_tutorial.html}{PyTorch library}), to build a sequential model alternating linear layers (14, 1000) $\to$ (1000, 230) $\to$ (230, 2) and ReLU layers to break linearity.
    \item Logistic regression (using the \href{https://scikit-learn.org/stable/modules/generated/sklearn.linear_model.LogisticRegression.html}{Scikit-learn library}, with the ``liblinear" solver, performing approximate minimization along coordinate directions) -- to introduce a simpler model in the benchmark.
\end{itemize}

For each type of model, a baseline model was trained with these default parameters, regardless of fairness objectives. Then to investigate the convergence of \fairdream\ towards Equalized Odds, we analyzed the model through the lens of Demographic Parity, as if Demographic Parity was the initial fairness purpose set by lawmakers. For this experiment, we the gap ratio to 3, meaning that when inequalities in OPR exceeded 3:1 (e.g. eligible men for a loan = 42\% versus 11\% for women), we started a correction to mitigate gaps created by the model on that feature (e.g. sex). 

With the same default parameters as the baseline model, \GridSearch\ and \fairdream\ tested new weights on 10 new models. 
As before, the goal of the competing models was therefore to simultaneously maximize a global statistical criterion (\hyperlink{https://scikit-learn.org/stable/modules/generated/Scikit-learn.metrics.roc_auc_score.html}{ROC-AUC}) and equalize the overall positive rates across groups (Demographic Parity). 

\subsection{Results}

Our hypothesis is that if we ask \fairdream\ to aim towards a fairness objective which is not conditional on true labels (in this experiment Demographic Parity), \fairdream\ would nevertheless tend to equalize the groups according to a fairness measure conditioned on true labels (here Equalized Odds). By comparing the best model of \fairdream\ and the best model of \GridSearch\ were selected, we found confirming evidence for our hypothesis. 

The full results of our experiment are accessible in the GitHub benchmark repository: \url{https://anonymous.4open.science/r/weights_distortion_impact-15/}.
Here, we present those that concern the objective of narrowing the gap in OPR between the advantaged and the disadvantaged group (Demographic Parity). Each column corresponds to an indication of performance of the two methods across the various classifiers chosen (except neural nets, on which FairDream and GridSearch implemented random classifiers to reach perfect Demographic Parity in both cases, yielding no difference). In total, across the XGBoost, logistic regression, and random forest classifiers, 16 corrections took place in both models. For the five metrics chosen (OPR, FPR, TPR, ROC-AUC-PR-AUC),\footnote{The table below leaves aside a sixth metric, of calibration. We refer to the online repository for the inclusion of that metric, and to the appendix for the definition of calibration.} each cell indicates, of these 16 corrections, how many times a model left a disparity between the best-treated and the worst-treated group that was 5\% or greater than the disparity left in the concurrent model relative to that metric. {For PR-AUC, the two numbers do not add up to 16 because in two cases the differences were less than 5\%.} Thus, a lower number indicates better performance of a model compared to the other.


\begin{table}[h]
\caption{Results of the benchmark of \GridSearch\ and \fairdream\ to equalize overall positive rates (OPR), with true positive rates (TPR) false positive rates (FPR), AUCs (ROC and PR)
}
\label{table:benchmark_grid_search_fairdream}\medskip
    \centering
    \begin{tabular}{|c|c|c|c|c|c|c|c|}
    \hline
        \scriptsize \textbf{Max Gap between groups} & 
 \scriptsize \textbf{OPR} & \scriptsize \textbf{FPR} &\scriptsize \textbf{TPR} &\scriptsize \textbf{ROC-AUC} &\scriptsize \textbf{PR-AUC} 
 &\scriptsize \textbf{Total} \\ \hline
        \GridSearch & \textbf{7} & 9 & 10 & 11 & 12 
        & 49 \\ \hline
        \fairdream & 9 & \textbf{7} & \textbf{6} & \textbf{3} & \textbf{2} 
        & \textbf{27} \\ \hline
    \end{tabular}
\end{table}

The table shows that \GridSearch\ slightly outperformed \fairdream\ with regard to the objective of Demographic Parity (i.e. OPR), namely in 9 out of 16 cases. However, this is at the cost of a decrease in predictive performance. On both AUCs, \GridSearch\ is less accurate. 
A closer look at the data also reveals that the mean ROC-AUC of all \GridSearch\ models is 59\%, which is a loss of 20\% compared to the baseline classifier (79\%). In all cases, Demographic Parity is achieved by increasing true positive rates, but thereby wrongly enhancing the false positive rates. On the other hand, \fairdream\ gets better results with regard to the objective of Equalizing Odds (i.e. TPR and FPR), with a ROC-AUC that remains high ($78\% \mp 1\% $ compared with the baseline model). 



To give further insights on the results, let us consider the comparison between the \GridSearch\ and \fairdream\ in their effort to mitigate the gaps in OPR between men and women generated by a baseline random forest. 
After correction, we see that \GridSearch\ has drastically modified the OPR of women (now 100\% are predicted to earn over \$ 50,000,compared to less than 5\% in the baseline model), and also increased that of men. \fairdream\ is more conservative (Cf. Figure \ref{fig:random_forest_sex_demographic_parity}), since the OPR for men is nearly the same, but the OPR for women is raised to 20\%. 

\begin{figure}[h]
\caption{Evaluation on Demographic Parity - Overall Positive Rates with \fairdream, \GridSearch\ and Baseline Random Forest Models on Sex}\label{fig:random_forest_sex_demographic_parity}
\centering
\includegraphics[scale=0.5]{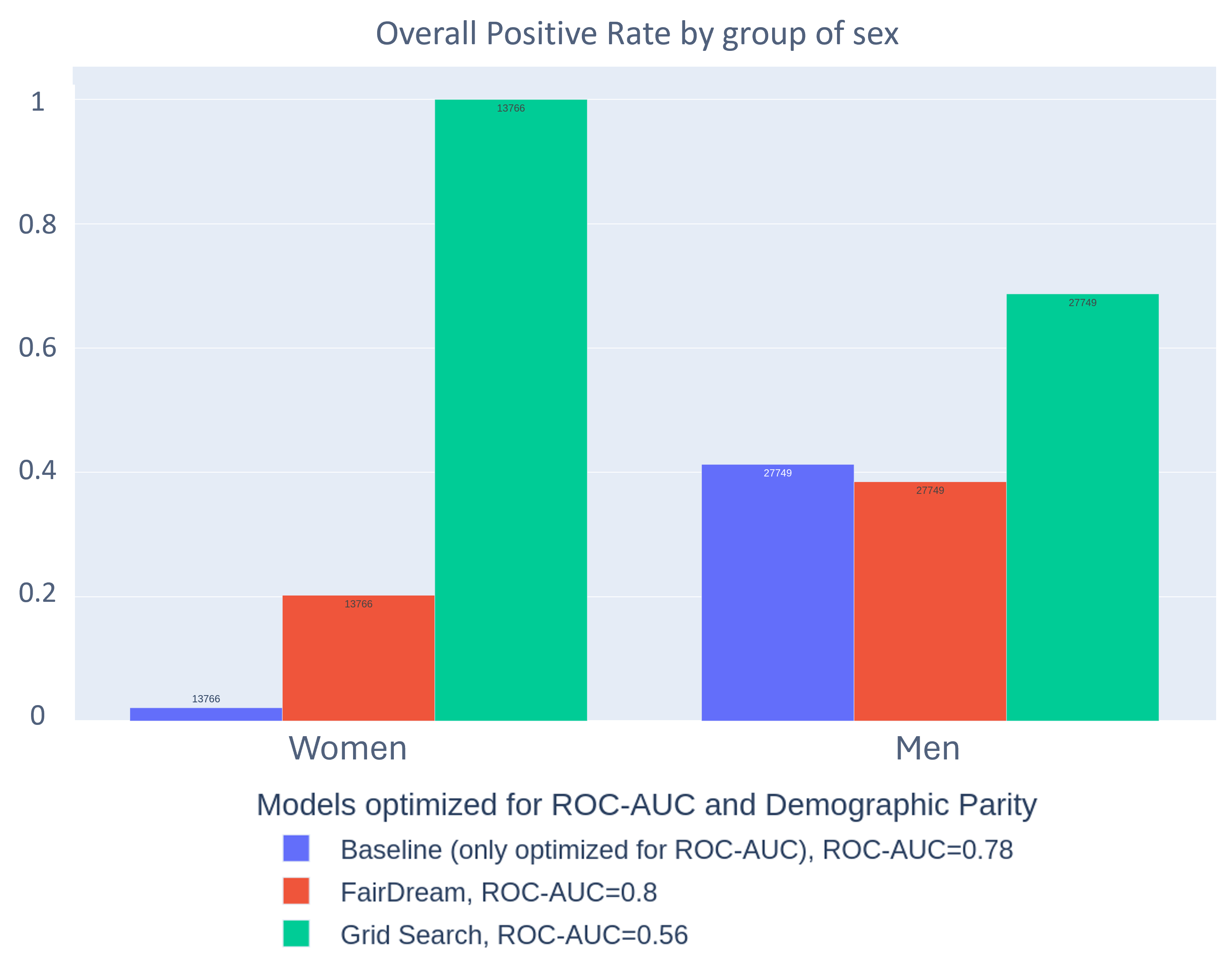}
\end{figure}

\begin{figure}[h]
\caption{Evaluation on Equalized Odds - True and False Positive Rates with \fairdream, \GridSearch, and Baseline Random Forest Models on Sex}\label{fig:random_forest_sex_equalized_odds}
\centering
\includegraphics[scale=0.5]{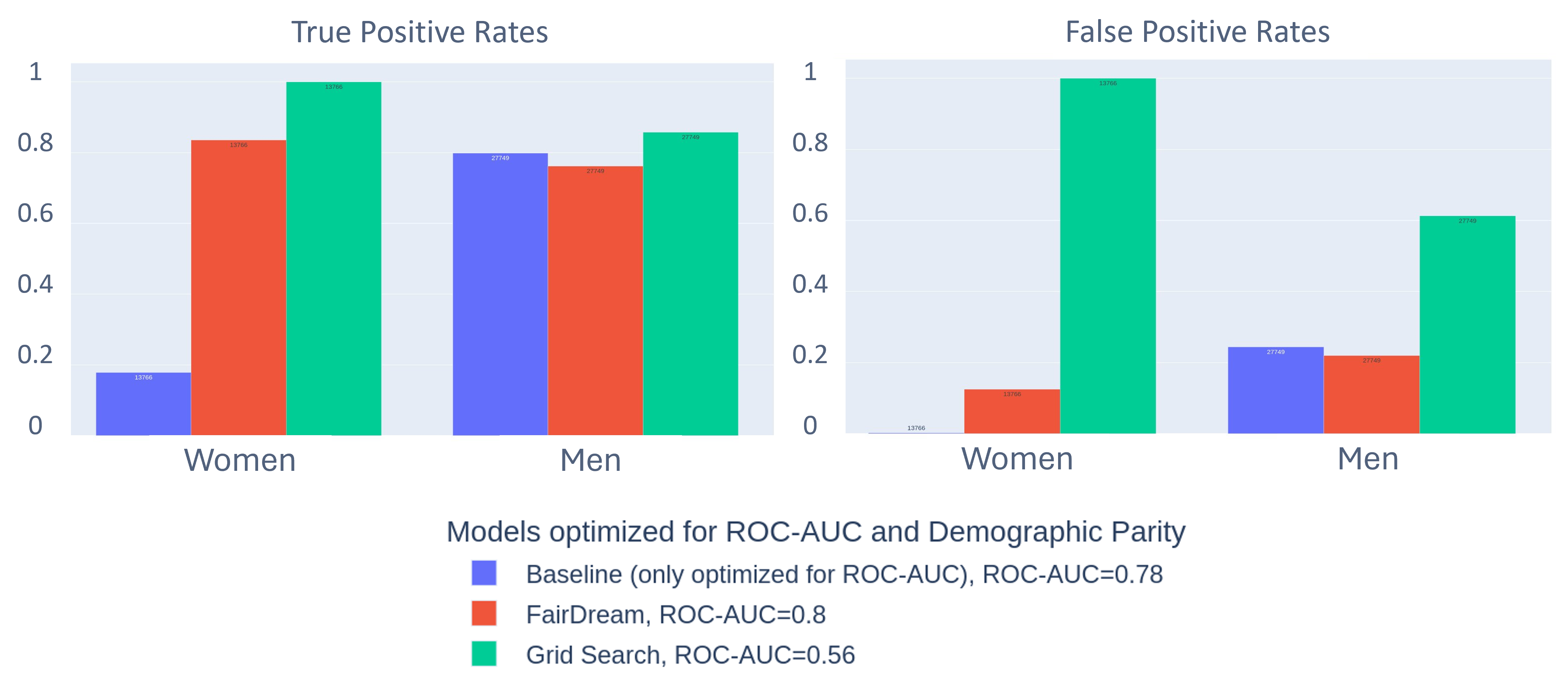}
\end{figure}

When we consider true positive rates, we get a different picture (Cf. Figure \ref{fig:random_forest_sex_equalized_odds}, left). In \fairdream, the percentage of individuals correctly predicted to earn over \$50,000 is now closer between the sexes (83\% vs 76\%, now slightly promoting women, where this difference is of 18\% vs 80\% in the baseline model). Likewise, false positive rates between men and women come out closer than in the baseline model by \fairdream\ (Cf. Figure \ref{fig:random_forest_sex_equalized_odds}, right: 13\% vs 22\%, where this difference is of 0.2\% vs 24\% in the baseline model). So \fairdream\ systematically responds to the objective of Demographic Parity by narrowing the gap in OPR between men and women, but it does so in a way that complies with the criterion of Equalizing Odds.




By increasing the OPR of women to 100\%, \GridSearch\ behaves almost like a random classifier (ROC-AUC = 56\%). By contrast, \fairdream\ narrows the gap between the sexes by improving the accuracy of the baseline classifier (ROC-AUC = 80\%). Notably, we see that neither \GridSearch\ nor \fairdream\ perfectly succeeds to enforce Demographic Parity in this example. \GridSearch's corrections are more drastic and do more to narrow the gap between the sexes, but at the expense of statistical accuracy.




\subsection{
Reweighting {versus} Optimization under Constraint}\label{subsection:explain_deviation}



How can we explain the difference in behavior between \GridSearch\ and \fairdream? The answer is that they implement the fairness objective in different ways.

- In \ref{eqn:grid_search}, the fairness objective $\mathcal{F}\left(\phi_{\theta,n} \right)$ is a new term added inside the global loss function (this way of correcting is referred to as an optimization of model's parameters \textit{subject to a fairness constraint} $\mathcal{F}$, see \cite{wan2023processing}). The left part of $L$ is a classic loss function, increasing when the error between the prediction $\hat{Y_i}$ and the true $Y_i$ differs. But the right part is a fairness loss function, added to the standard measurement of error, increasing when the fairness constraint $\mathcal{F}$ is violated. To achieve Demographic Parity, the fairness term of the loss incentivizes the model to predict $\hat{Y_i}=1$, even if the true label of a 17-29 year-old individual $i$ is $Y_i=0$.

- \fairdream\  has no such internal constraint inside its cost function $L(\phi_{\theta,n} ,\ W_n)$. 
To equalize the percentage of individuals predicted as earning over \$50,000, which is the purpose of Demographic Parity, the harmed individuals of the group $S_k$ are provided with a new weight of error $w_n(S_k)$, growing with their previous disadvantage. However, once the new sample weights are computed, they become coefficients of the classic loss function in \ref{eqn:fairdream}.

For example, the new weight $w_2(S_1)=0.8$ of any 17-29-year old individual means that if a 17-29-year old truly has the property of earning less than \$50,000 ($Y_i=0$), predicting them as earning over \$50,000 ($\hat{Y_i}=1$) amounts to 8 times the cost of any prediction error for 29-37 years old (weight of 0.1). The \fairdream-style classifier is, therefore, strongly incentivized to give predictions in accordance with the true label $Y_i \in \{0,1\}$. 
This means that even though \fairdream\ does its work to narrow the gap in overall positive rates, it does it subject to the underlying constraint of equalizing odds.


\section{Choosing a Fairness Metric conditioned on True Labels}\label{section:decision_rule}

We provided a computational and mathematical explanation for the deviation between the initial fairness criterion (Demographic Parity) and the outcome actually achieved by \fairdream\ (Equalized Odds). In this section and the next one, we move to a normative discussion of the choice between these two fairness criteria. Here, we first discuss whether Equalized Odds is too conservative from the outer perspective of a policy maker. In the next section, we will examine the constraint from an inner perspective, this time from the point of view of an individual in the situation of applying for a loan.

\subsection{Fairness from Accuracy
}\label{subsection:normative_claim_income}

In our benchmark comparison, each time the \GridSearch\ models outperformed \fairdream\ with respect to Demographic Parity, it was at the cost of a worse statistical performance.
By contrast, \fairdream\ gives a more accurate picture of the ground truth:
the \fairdream\ correction method produces a low rate of false positives (less than 5\%) and a high rate of true positives (nearly 90\%). This means that \fairdream\ is a method that increases the Sensitivity (i.e. TPR) and the Selectivity (i.e 1-FPR) of the baseline model on which it operates.
The question raised by this situation is whether increasing statistical accuracy in the sense of Selectivity and Sensitivity is automatically a way of increasing fairness. 

Several arguments can be given against the sufficiency of accuracy to achieve fairness. The main one is that even the most accurate classification may simply replicate imbalances that are in the data ahead of the algorithm's workings, as a result of social biases or social injustice. We agree with this argument, and we grant that improving descriptive accuracy may not be \textit{sufficient} to achieve normative fairness (in line with Hume's classic remarks on the is-ought distinction \cite{hume1958treaty}).
However, we consider that descriptive accuracy in predictions should at least be viewed as a \textit{necessary} condition on fairness \cite{wachter2020bias}, and that it may even be a necessary and sufficient condition in cases where the predictions themselves concern matters of fact. 

In such cases, the case for conditional fairness metrics can be supported by general epistemological considerations. A user trying to narrow the gap between overall positive rates predicted by our \fairdream\ classifier (16\% for 17-29 years old and 59\% for 29-37 years old, Cf. Table \ref{fig:best}) may be viewed as falling prey to a ``base rate neglect" (\cite{tversky1982evidential}). 
In the present case, the base rate, or the ratio of actual positives to the whole population, is radically different in each age group. Less than one in six young people actually earn \$50,000, while four in six older people actually earn \$50,000.\footnote{Base rates are 547/3616=15\% for 17-29-year-olds, versus 2400/3666=65\% for 29-37-year-olds (see Table \ref{fig:best}).} Assuming the data is reliable, it would not be possible for a classifier to predict more than 15\% of the 17-29 and 65\% of the 29-37 to earn \$50,000 without increasing false positives (thereby decreasing in predictive performance). 



The main issue with increasing false positives concern downstream effects (see \cite{barocas2023fairness}): blindly enforcing Demographic Parity by increasing the false positive rate in the younger group is also susceptible to lead to harmful consequences. Ceteris paribus, it can put the younger group at higher risk of defaulting their loan, assuming income is causally the main variable to sustain loan refunding, and that the algorithm's prediction on income is critical in deciding whether or not to grant credit.


In the case of \fairdream, it may be added that the method's reliance on Equalized Odds is not as conservative as it might seem. In particular, it may be objected that if Sensitivity and Specificity are key here, then they should be plotted for each decision threshold, which is what ROC-AUC represents. Our initial example is reminiscent here of the debate between ProPublica and Northpointe: both protagonists, after all, agree that both Selectivity and Sensitivity matter for fairness; but Northpointe's claim is that the tradeoff between them should be measured at all thresholds by comparing the ROC-AUC, rather than by looking at just one threshold. Despite that, Northpointe also grants that given a particular decision boundary, one may compare true positive rates and false positive rates between groups to check for inequalities in accuracy. Their proposed criterion is that if the ratios between TPR is less than 1 at some threshold, and the ratio of FPR is greater than 1, then one can conclude for inequality, but they concede that in cases in which only one of these two conditions is satisfied, then cost considerations may be brought in.\footnote{See \cite[p. 16]{dieterich2016compas}: ``The
[Ratio of FPR] is less than 1, but the [Ratio of TPR] is not greater than 1. This is an inconclusive
result. When the results of a comparison of the accuracy of a binary test
in two groups is inconclusive, the results can be consolidated using a cost
function (Pepe, 2003). That work is beyond the scope our report''.} The situation depicted in Table \ref{fig:baseline} has this property: there, the imbalance exceeds 1 only in the false positive rate. However, it does so by a factor of 10 (0.2 in the 29-37 group compared to 0.02 in the 17-29 group). Even for those who maintain that COMPAS is not biased, this is a case where the imbalance justifies intervention.




As shown in Table \ref{fig:best}, the proposed intervention by \fairdream\ does narrow the gap in Demographic parity. The fact that it does this cautiously, namely within the bounds of Equalized Odds, still corresponds to a policy that is less conservative than one that would suggest a revision only in cases in which the ratios of TPR and FPR both exceed the threshold of 1 discussed by Northpointe in particular.

\subsection{Always prefer Equalized Odds?}
\label{subsection:trade_off}

The foregoing arguments notwithstanding, there remain reasons to doubt the validity of conditional metrics in every situation. {In order to answer this worry, we use the same methodology, namely we need to vary the case and to consider more examples.} 
For comparison, consider another realistic financial classification task in which the bank-teller has to estimate which client is a “good risk” or “bad risk” based on age, sex, profession, accounts, and credit amount, as appears in the \href{https://www.kaggle.com/datasets/uciml/german-credit}{German Credit dataset}.\footnote{\url{https://www.kaggle.com/datasets/uciml/german-credit}.} 


Here, a hidden bias may very well have influenced the decision to label clients as ``good risk'' or ``bad risk'', simply because the labels ``good'' and ``bad'' are not purely factual (unlike ``earning over \$50,000''), but evaluative and judge-dependent (\cite{solt2018multidimensionality,icard2023measuring}).  The bias may be reflected in spurious correlations between variables, for instance between ``good risk'' and male.
Applied to this distribution of the labels, a model achieving Equalized Odds will predict overall positive rates which replicate wrong or undesirable base rates, under-representing the “good risk” women.

The choice between conditional and unconditional fairness criteria represents two different visions of distributive justice (\cite{rawls1999justice}). An appeal to unconditional fairness criteria is justified when the basic labels and their distribution can be suspected of an initial unwanted bias \cite{wachter2020bias}. 
On the other hand, the use of conditional metrics is justified when the goal is to ground prediction in descriptions and distributional properties of a population that need to be faithfully represented \cite{dwork2012fairness}.

In our running example of income prediction, the transparency of labels and their factual character is undeniable. The distribution of income levels, on the other hand, may be an object of deeper social debate (viz. should the youth's first income be raised? should the capacity to refund a loan be based on other indicators for this age group?).\footnote{See for example \cite{taylor2020junior}} But merely changing the labels would not appear to be an adequate or reliable way of affecting the underlying distribution of incomes. In a task involving the labels of “good” {versus} “bad” risk clients, on the other hand, those very labels are no longer transparent, and the various methods of correction we distinguished (pre-processing, in-processing, and post-processing) may be used to remove hidden biases. 

Acting on the labels is a delicate matter, which can be justified if the labels are evaluative, even implicitly, and if there are associated levers to enforce the new distribution of resources (public policies, internal rules, and means inside corporations). 
In particular, an understanding of causal relations is required to identify the features on which to correct the labels, in order not to harm new individuals. One should also carefully study the effects of changing the risk labels on the basis of features like income, or geographic area.\footnote{As a first step, the effects of overall fairness metrics on new allocation of outcomes could be validated with A/B testing on real populations (\cite{saint2020fairness}). Controlled experimentation on the effects of new labels distribution would defuse two problems of algorithmic fairness spotted by \cite{selbst2019fairness}, namely the ``portability trap'' and the ``ripple effect trap''.}


It is important to note that \fairdream\ was implemented to inform users of allocation disparities between various groups. 
It is possible, of course, to imagine cases where a gap in OPR is detected even as the TPR and the FPR are equal between two groups from the get go, hence in cases for which sensitivity and selectivity are actually perfect. 
For such cases, which imply different base rates between the groups, we are not saying that no intervention should ever be made. The Census case we started off with is of this kind: on a variant of it, each adult in their group is perfectly predicted for their income, and the difference in overall positives is merely due to the fact that, on average, younger adults earn less than the older cohorts. \fairdream\ in such cases will not offer to close this gap, but it will still be instrumental in providing information \cite{mayson2018bias} about inequalities that may be structural. 


\section{\pe{Conditional dependencies}}\label{sec:simpson}

%
%
\pe{Finally, in this section, we examine \fairdream's convergence towards Equalized Odds from the view point of an individual in the situation of being lent a loan.\footnote{We thank the editors for urging us to address this issue.}}
\pe{Has the \fairdream\ model — whose predictions are shown in Table \ref{fig:best}  — actually become \textit{fairer} than the initial model shown in Table \ref{fig:baseline}? We think the answer is positive, but it involves a careful look at dependencies between the relevant variables. The results produced by \fairdream\ bear a resemblance to Simpson's reversal paradox in that regard (\cite{pearl2009causality,fitelson2017confirmation}), and we propose to use it as a guide toward our conclusion.}

\pe{
Let us first establish the analogy with Simpson's paradox. On the one hand, the new model appears to strongly favor clients aged between 29 and 37 compared to the 16-29 group. On the other hand, when we split each group into those truly earning over \$50,000 and others, this relation reverses within each group. That is, between the two age groups, the probability of being classified as earning more than \$50,000 is equalized with \fairdream. Formally, we have the following relations (where ``$A=1$'' stands for membership in the 29-37 group; ``$B=1$'' membership in the 16-29; ``$Y=1$'' for truly earning over \$50,000, ``$\gg$'' expresses ``much greater than'', and ``$\approx$'' approximate equality):}
$$p(\hat{Y} = 1 \mid A = 1) \gg p(\hat{Y} = 1 \mid B = 1)$$
yet,
$$ 
p(\hat{Y} = 1 \mid A = 1 \ \wedge\ Y = 1) \approx 
p(\hat{Y} = 1 \mid B = 1 \ \wedge\ Y = 1)
$$
$$
p(\hat{Y} = 1 \mid A = 1 \ \wedge\ Y = 0) \approx 
p(\hat{Y} = 1 \mid B = 1 \ \wedge\ Y = 0)
$$

\pe{Technically, Simpson's paradox requires a \textit{strict} inversion 
{within} the partition groups (here, our groups of true labels).\footnote{We are indebted to Branden Fitelson for helping us clarify this matter.} 
For it to qualify as a genuine Simpson's paradox, it would have to be the case that,
for every true label $i \in \{0,1\}$,}
\[
p(\hat{Y} = 1 \mid A = 1 \ \wedge\ Y = i)
\;\leq\;
p(\hat{Y} = 1 \mid B = 1 \ \wedge\ Y = i).
\]
\pe{The direction of this inequality is not respected by the \fairdream\ model, since we have:}
\[
p(\hat{Y} = 1 \mid A = 1 \ \wedge\ Y = 1) = 88\% 
\;<\;
p(\hat{Y} = 1 \mid B = 1 \ \wedge\ Y = 1) = 89\%
\]
\[
p(\hat{Y} = 1 \mid A = 1 \ \wedge\ Y = 0) = 4\%
\;>\;
p(\hat{Y} = 1 \mid B = 1 \ \wedge\ Y = 0) = 3\%.
\]
\pe{Among the younger clients who were formerly disadvantaged, 
\fairdream\ now selects a slightly larger proportion of individuals 
than among their older counterparts within one group ($Y = 1$), 
but not within the other ($Y = 0$). Yet, this difference amounts to only 1\%. Because of that, the fairness intervention occasioned by \fairdream\ is a case of near-Simpson reversal, and we can draw on analyses of Simpson's paradox to tease apart two readings of the fairness intervention.}

\pe{Consider a 16-29 year-old who is wondering about their chances of being granted a loan, assuming it depends on an adequate recognition of their income. They learn that their chances (16\%) are significantly lower than for a 29-37 year old (59\%) even after correction by \fairdream. Prima facie, this seems to support the following conditional statement:}

\ex.\label{ex:strict} If I were in the 29-37 age group, my chances of getting a loan would be significantly higher than they are.

There is a reading of this conditional that makes it true, but it constitutes a weak case for lack of fairness. For the reading that makes this conditional true does not actually keep constant the person's income under the counterfactual supposition changing the person's age. In fact, what the conditional assumes implicitly is that if I were in the 29-37 age group, I would be more likely to be richer than by being in the 17-29 group. 


Our argument is that either of the following conditionals gives a much better evaluation for fairness or lack of fairness, depending on the person's income, which they are taken to know:

\ex.\label{ex:var1} If I were in the 29-37 age group, and I earned over \$50,000 (as I do), my chances of getting a loan would be significantly higher than they are.

\ex.\label{ex:var2} If I were in the 29-37 age group, and I earned less than \$50,000 (as I do), my chances of getting a loan would be significantly higher than they are.

\pe{Note that these two conditionals would come out true of the uncorrected model of Table \ref{fig:baseline}, but they come out false of the corrected model proposed by \fairdream. In other words, \fairdream\ achieves near conditional independence of the predictor's decision relative to the age variable, unlike the baseline model.}

Using $a$ for ``I am in the 29-37 age group", $b$ for ``I earn over \$50,000'' and $c$ for ``my chances of getting a loan are higher than  they are'', we can represent the previous conditional sentences more succinctly as in the following table (we use $\boxright$ to represent the counterfactual conditional, and assume that $\wedge$ binds tighter than $\boxright$). In each column, we have indicated the truth values compatible with each sentence in the context. For $a\boxright c$, ``False / True'' is to indicate that the sentence admits two different interpretations leading to different judgments. 


\[
\begin{tabular}{cr|c|c}

& Sentence & Baseline model & \fairdream\ model\\

\hline

\ref{ex:strict} & $a\boxright c$ & True & False / True \\

\ref{ex:var1} & $a \wedge b \boxright c$ & True & False\\

\ref{ex:var2} & $a \wedge \neg b \boxright c $ & True & False\\

\end{tabular}
\]


 For the \fairdream\ model, the  triad of judgments $\langle$True, False, False$\rangle$ is actually inconsistent relative to standard treatments of counterfactual reasoning. In Stalnaker's or Lewis's semantics for counterfactual conditionals (\cite{stalnaker1968conditionals,lewis1973counterfactuals}), the sentence $a\boxright c$ entails the disjunction $ (a\wedge b \boxright c) \vee (a\wedge \neg b \boxright c)$ and conversely, so  the falsity of both disjuncts must entail the falsity of $a\boxright c$. The reason is that the closest world (or worlds) where I am in the 29-37 age group is either a world where I earn over \$50,000, or a world where I earn under \$50,000. Which world it is is determined by which of these two options holds in the actual world. 
 Our claim is that the interpretation under which \ref{ex:strict} is false is precisely the one that underlies the falsity of \ref{ex:var1} and \ref{ex:var2} relative to \fairdream\, and that is expressed probabilisticaly in terms of Equalized Odds.

  By contrast, the reading of \ref{ex:strict} on which it is true corresponds to a counterfactual assumption in which varying the age involves an adjustment of one's actual income. This is akin to a backtracking reading of the counterfactual, one in which, in order to satisfy the antecedent, some additional adjustments need to be made (\cite{lewis1979counterfactual,khoo2017backtracking}). But this no longer corresponds to a ceteris paribus or minimal revision of one's current situation.\footnote{For an explicit formalization of ceteris paribus reasoning with counterfactuals, see \cite{girard2018prioritised}.} A more adequate paraphrase for the intended reading of \ref{ex:strict} would therefore be: ``if I were a person drawn randomly from the 29-37 age group, then on average I would stand higher chances of getting a loan that if I were a person drawn randomly from the 16-29 age group''. This is true, but this is not an adequate basis to conclude that the intervention produced by \fairdream\ would be biased or unfair in how it treats different groups. For the base rates already show a similar imbalance regarding income across the groups, prior to any intervention. As explained previously, this argument assumes that this lower base rate is not itself a case of structural injustice. Structural injustice would support the true reading of \ref{ex:strict}, but would not count as evidence that the algorithm is unfair.

\pe{In summary, we thus take the correction at play in \fairdream\ to implement a notion of fairness that comes closer to ceteris paribus reasoning than the \GridSearch\ methods we discussed. 
This is, ultimately, why we think of Equalized Odds as encapsulating a more adequate notion of fairness than Demographic Parity}.

\section{Conclusion}\label{sec:conclusion} 

In this paper we presented an algorithmic fairness tool to detect and correct for disparities in classification, the package \fairdream. The practical goal of this package 
is to help lay users implement their vision of fairness, in order to come up with socially optimal decisions. Our study  of the specific properties of \fairdream\ raised an explainability issue: even as the user sets as fairness goal to narrow the gap in overall positive rates between different groups, the correction does not achieve Demographic Parity, but instead appeared to be constrained by Equalized Odds.






In the first part of this paper, we provided an in-depth explanation of this specific property of \fairdream. We conducted a systematic benchmark to compare \fairdream\ with the state-of-the-art correction method \GridSearch, testing the two methods on different features and machine-learning models. First, we proposed an explanation for this difference by looking at the specifics of each algorithm. \GridSearch\ incorporates the fairness constraint as an additional cost inside the cost function of the model. Due to the fairness cost, this family of methods is more likely to deviate from the true labels (increasing false positives), in order to achieve Demographic Parity. On the other hand, \fairdream\ introduces a differentiated attention to harmed individuals within the cost function of the model. Therefore, it increases accuracy on harmed groups (through enhanced true positives). By basing fairness on accuracy, \fairdream-type methods are thus bound to make the overall positive rates converge towards their base rates. As a result, they will not necessarily achieve Demographic Parity. 


Our study of \fairdream\ led us to 
a normative claim about how to approach fairness.
In the second part of the paper we provided a normative justification for \fairdream's correction method.
By choosing a metric conditional on true labels, the user implicitly commits to the idea that there is sufficient agreement on the transparency of labels and on the shape of their distribution. By selecting an unconditional metric, one commits to the idea that the labels themselves may have introduced unwanted disadvantages, 
or that the input distribution must be corrected rather than replicated. In the case of \fairdream, the correction method may be criticized for being overly prudent, but as we argued, more revisionary methods must be carefully justified and cannot be applied lightly. Finally, we have established a parallel between the correction operated by \fairdream\ and Simpson's paradox to argue that Equalized Odds expresses a more constrained notion of counterfactual evaluation of fairness than does Demographic Parity.

\pe{Our investigation of \fairdream\ allows us to draw a more general lesson concerning the explainability of AI models. Even as the user selects the overall positive rates as a diagnostic criterion for inequality between groups, the correction for that gap is subject to more constraints.
As we have argued, the transparent use of a fairness intervention algorithm involves an explanation of the reweighting method used by the algorithm, a contrastive look at nearby systems governed by similar correction mechanisms that produce different outcomes, and a normative justification of the dependence of the correction methods on true labels.}


We conclude that AI classification systems should preferably be released with an indication of these features: not only an indication of the direction they intend for the correction method, but also an indication of how they actually perform relative to this goal, of how they perform on alternative datasets, and relative to minimally modified variants on the same data. Making this information available eliminates arbitrariness in classification and is a step toward more trust, transparency, and control on the side of lay users}. 

\bibliographystyle{apalike}
\bibliography{references.bib}

\begin{appendix}
\section*{Appendix: Calibration and Precision}
\addcontentsline{toc}{section}{Appendices}
\renewcommand{\thesubsection}{\Alph{subsection}}
\ts{Besides Demographic Parity, the criterion of Equalized Odds is also commonly opposed to various notions of predictive parity, understood either in terms of equal calibration (\cite{pleiss2017fairness,barocas2023fairness}), or in terms of equal precision relative to a threshold (\cite{dieterich2016compas,mayson2018bias}). In this appendix, we start with some data relative to \fairdream's performance on calibration, and we explain why, on a normative basis, calibration for us does not constitute a relevant criterion.}

One common view of fairness is found in the idea of \textit{predictive parity} between groups. Predictive parity can mean that the same fraction of true positives is predicted between groups at identical risk scores, or it can mean that the same fraction of true positives is predicted between groups relative to a classification threshold. The first notion corresponds to the statistical notion of \textit{calibration} of a classifier, whereas the second corresponds to the notion of statistical \textit{precision} of a classifier. Both notions are sometimes indistinctly referred to under the term ``calibration'' in the literature (see \cite{mayson2018bias,long2021fairness}), although strictly speaking they differ.\footnote{Both notions have also been invoked in defense of COMPAS against ProPublica's charges of racial bias. In their response to ProPublica, Northpointe argued that at the cutoff chosen by ProPublica to show Unequal Odds between Whites and Blacks, the statistical precision of the COMPAS algorithm was roughly the same between both groups. Moreover, it was shown that at each risk score, the proportion of rearrest between groups was roughly the same (see \cite[Figure 2]{corbett2017algorithmic}), {\cite{dieterich2016compas} produce very closely related data (see their tables A1 and A2), but they look at predictions \textit{above} all possible risk scores, which is not strictly speaking calibration.}}



Both calibration and precision are central notions in that regard, and one may wonder how \fairdream\ fares with regard to either. Let us consider calibration first. Calibration can be looked at either in a relative, or in an absolute sense (\cite{eva2022algorithmic}). In a relative sense, it requires that for every possible risk score $R$, the proportion of individuals assigned that risk score that actually have the predicted property be equal between groups. We say that an algorithm satisfies \textbf{Equal Calibration} between two groups $A$ and $B$ with respect to $Y$ provided for every value $r$ of its risk score $R$: 
   $$p(Y=1 |R=r\wedge A=1)=p(Y=1|R=r\wedge B=1)$$ 
   
We select this relative sense as it is the one commonly adopted to link fairness and calibration (see \cite{corbett2017algorithmic,hedden2021statistical,barocas2023fairness}). 
A distinct, non-relative sense of calibration, requires that the method used to classify be itself well-calibrated \textit{within} a given group this time, that is, that for each possible risk score $r$, the ratio of people classified as positive who are actually positive be equal to $r$, or sufficiently close to $r$ (\cite{kleinberg2016inherent}). This is a stronger requirement, for a method may fail to be well-calibrated in that absolute sense, but still be such that it selects the same percentage of true cases in different groups.

\begin{figure}[t]
\caption{Calibration Curves of \fairdream\ and Baseline Random Forest Models on Sex.}
\label{fig:rf_sex_calibration_curves}
\centering
\includegraphics[scale=0.5]{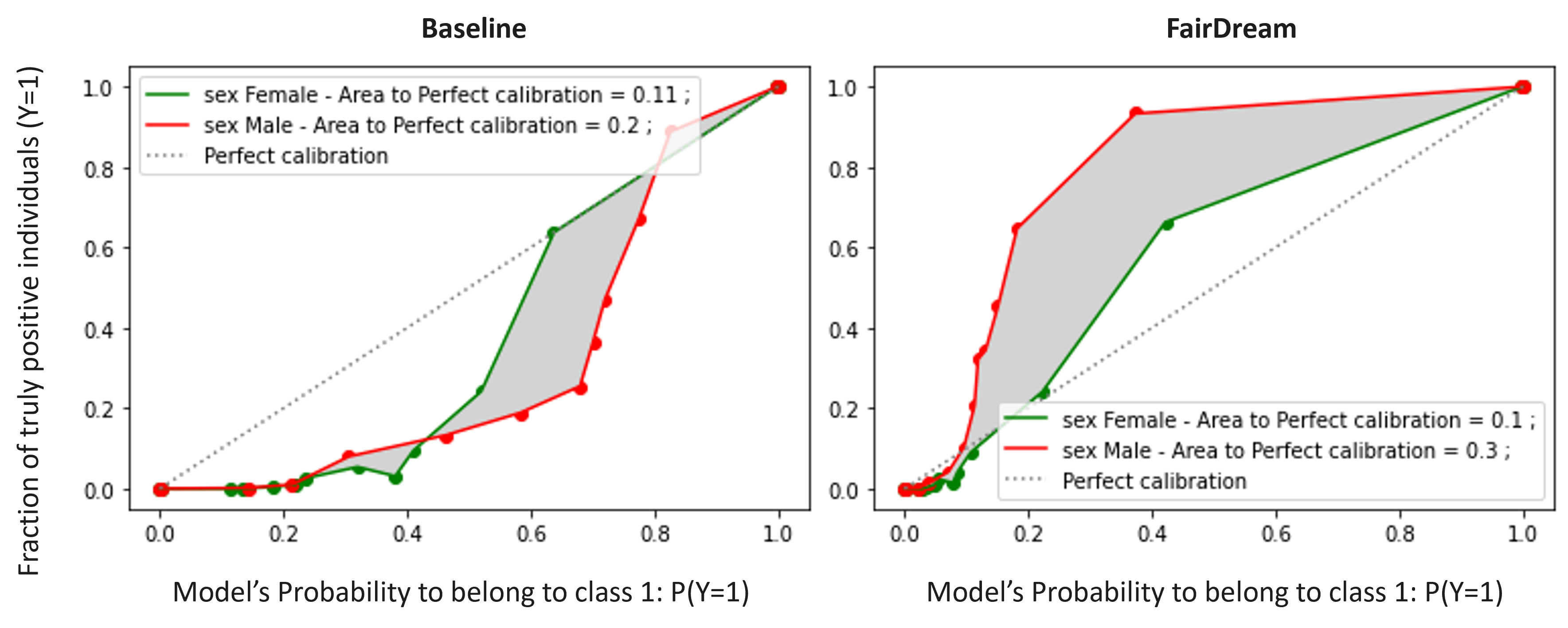}
\end{figure}

It is a fact that in general, achieving Equalized Odds implies sacrificing Equal Calibration (see \cite{kleinberg2016inherent},\cite{barocas2023fairness}). The case of the random forest models used above illustrates this trade-off. When we compare and sum the scores (here, probabilities of getting $Y=1$), the calibration gap between groups is increased by \fairdream\ compared to the Baseline model. The area between calibration curves, approximated through a trapezoidal rule, is larger for \fairdream\ than for the initial model (0.2 versus 0.09, see Figure \ref{fig:rf_sex_calibration_curves}). 
That \fairdream\ favors Equalized Odds at the expense of Equal Calibration is also confirmed in the other cases. Out of 16 features where correction happened, \fairdream\ reached the highest gaps between groups in 11 calibration curves (see Table \ref{table:benchmark_baseline_fairdream}.)


\begin{table}[!ht]
\caption{Comparison between Baseline and \fairdream\ on benchmark: each cell reports the number of times a model achieves the highest gap between groups on a metric.
}\label{table:benchmark_baseline_fairdream}\medskip
    \centering
    \begin{tabular}{|c|c|c|c|c|c|c|c|}
    \hline
        \textbf{{\scriptsize Max Gap between groups}}  & \textbf{{\scriptsize OPR}} & \textbf{{\scriptsize FPR}} & \textbf{{\scriptsize TPR}} &  \textbf{{\scriptsize ROC-AUC}} & \textbf{{\scriptsize PR-AUC}} & \textbf{{\scriptsize Calibration}} & \textbf{{\scriptsize All metrics}}\\ \hline Baseline & 9 & 10 & 14 & \textbf{3} & 13 & \textbf{5} & 54\\ \hline
        \fairdream & \textbf{7} & \textbf{6} & \textbf{2} & 11 & \textbf{1} & 11 & \textbf{38}\\ \hline
    \end{tabular}
\end{table}

Figure \ref{fig:rf_sex_calibration_curves} also tells us something about absolute calibration, so on the positions of curves relative to the ideal calibration curve $x=y$. In the Baseline model, both curves are below ideal calibration:
$P(\hat{Y}=1)$ under-estimates the real percentage of women and men which truly earn over \$50,000. Whereas \fairdream\ reverses this tendency: individuals of both groups are over-estimated by the model. However, while switching from under- to over-estimation, \fairdream\ produces a better calibrated curve for women than men, this time judging by how much the female group's curve departs from absolute calibration. That is, while \fairdream\ undeniably widens the gaps between women and men to fulfill Equalized Odds, the situation of women is enhanced not only on true positive rates, but even on the calibration picture. Overall, therefore, neither the base model nor \fairdream\ is well calibrated in an absolute sense, but they err in opposite ways. Moreover, while the calibration gap is increased with \fairdream, this is a case where absolute calibration in the discriminated group is improved in \fairdream. 


We do not see Equal Calibration to be fundamental as a fairness constraint, however, and we do not see it on a par with Equalized Odds either. 
Indeed, like Demographic Parity, Equalized Odds assumes that a threshold has already been applied to the scores. 
The decision-maker is only able to act once this threshold has been set.\footnote{Like \cite{grant2023equalized}, we thus favor decision procedures over predictive methods to investigate the fairness of algorithm. Our defense of Equalized Odds differs from his, but both share this normative consideration.} 


Yet, as we see in Figure \ref{fig:rf_sex_calibration_curves}, calibration compares the model's properties along all scores, before any threshold has been set. As argued by \cite{corbett2017algorithmic}, moreover, one can find cases of Equal Calibration between groups for which a decision threshold is nonetheless prone to induce a differential treatment between groups, if the two groups' risk scores overlap on only part of the scale. 

A way to take into account decision thresholds is to compare Equalized Odds with a constraint of {Equal Precision} between groups, relative to the same threshold, that is,
An algorithm satisfies \textbf{Equal Precision} between two groups $A$ and $B$ with respect to $Y$ provided
$$p(Y=1|\hat{Y}=1\wedge A=1)=p(Y=1|\hat{Y}=1\wedge B=1)$$

Precision, namely the rate of true positives among all predicted positives (see Section \ref{section:definitions}), is a metric that Northpointe opposed to COMPAS in their rejoinder (\cite{dieterich2016compas}). More generally, \cite{long2021fairness} has argued that Equal Precision ought to stand as a necessary condition on fairness. However, we find it hard to endorse this generalization for all cases. For instance, to use an example of the same kind used by Long, we can come up with cases in which Precision is not equal between two groups, but Sensitivity (= True Positive Rate, aka. Recall) and Selectivity (=True Negative Rate) are identical, and for which we do not have the intuition that there is an unfair treatment. 

An example is given in Table \ref{tab:unequalprec}. Group 1 has 28 students, and Group 2 has 40 students, comprised of students deserving a High Grade and students deserving a Low Grade (where we assume these qualities to be objectively measurable). In both groups, the True Positive Rate and the True Negative Rate are the same: Group 1 and Group 2 have the same absolute number of true High Grade students, but in Group 2 we get more Low grade students wrongly classified as ``High'' than in Group 1. The Precision on the ``High'' label is only 1/7 compared to 1/5 in the first group. Since Selectivity and Sensitivity are identical in this example, this is, however, a case in which the criterion of Equalized Odds is satisfied.

\begin{table}[t]
\caption{Unequal Precision but equal Sensitivity and Selectivity ($Y=1$ means that the student's type is High, $Y=0$ that is is Low)}\label{tab:unequalprec}
\[
\begin{tabular}{c|c|c}

Group 1 & $\hat{Y}=1$ & $\hat{Y}=0$\\
\hline
$Y=1$ &  1 & 3 \\
\hline
$Y$=0 & 4 & 20 \\
 
\end{tabular}
\qquad
\begin{tabular}{c|c|c}
Group 2 & $\hat{Y}=1$ & $\hat{Y}=0$\\
\hline
$Y=1$ &  1 & 3\\
\hline
$Y=0$ &  6 & 30 \\
  \end{tabular}
  \]
\end{table}

Against sufficiency this time, we can easily find cases in which Precision on one label will be identical across groups, but for which we would have the intuition of the classifier being unfair or biased. Imagine a teacher having to grade only excellent, High profile students, male and female, in a testing experiment in which the teacher can issue either ``High'' or ``Low'' judgments. Suppose that the teacher always grades true High type male students perfectly as ``High'', but classifies only 1 in 5 female students as ``High'' and the others as ``Low'' (see Table \ref{tab:equalprec}). The precision on the ``High'' predictions is 100 percent in both groups, but obviously the Sensitivity is not the same for males and for females. This is a clear case of a biased judgment, where so many misses in the female group imply an unfair outcome.

\begin{table}[h]
\caption{Equal Precision, Unequal Sensitivity}\label{tab:equalprec}
\[
\begin{tabular}{c|c|c}

Males & $\hat{Y}=1$ & $\hat{Y}=0$\\
\hline
$Y=1$ &  25 & 0 \\
\hline
$Y$=0 & 0 & 0 \\
 
\end{tabular}
\qquad
\begin{tabular}{c|c|c}
Females & $\hat{Y}=1$ & $\hat{Y}=0$\\
\hline
$Y=1$ &  5 & 20\\
\hline
$Y=0$ &  0 & 0 \\
  \end{tabular}
  \]
\end{table}

The upshot is that Equal Precision is neither a necessary, nor a sufficient condition on fairness. In the case we just presented, one might respond that Precision in the ``Low'' judgments is not constant across the groups, even as precision on the ``High'' judgments is. But the problem is that to ask for Equal Precision on both labels is tantamount to asking for Equalized Odds, which would undercut Long's argument. 
This is not to say that Precision does not matter to \fairdream. As explained (see fn. \ref{fn:f1}), we conducted our benchmark so as to maximize the trade-off between Precision and Recall. From Table \ref{table:benchmark_baseline_fairdream}, we also see that the gap in PR-AUC diminishes in \fairdream\ across groups compared to the baseline. But since this is not a criterion we set a priori, no more than ROC-AUC, we do not build any conclusion on this fact. 

In summary, therefore, whether on a descriptive basis, or on a normative basis, Equal Calibration does not end up as a decisive criterion in our evaluation of \fairdream's correction, and neither does Equal Precision. 

\end{appendix}



\section*{Acknowledgements}

We express our deepest gratitude to Nicolas Meric and Johnathan Nguyen for their unfailing support throughout this project, as well as to the company DreamQuark, which funded Thomas Souverain’s doctoral research CIFRE at ENS-PSL under the leadership of its CEO Nicolas Meric. Without the participation of DreamQuark to the VERITAS hackathon, and the training received by Thomas Souverain in machine learning at DreamQuark, the package \fairdream\ would not have come into being.

We acknowledge the programs ANR TRUSTEDNEWS (ANR-25-ASM2-0003-01), Program ANR HYBRINFOX (ANR-21-ASIA-0003-04), Program ANR FrontCog (17-EURE-0017), the Agence nationale de la recherche technologique (Thomas Souverain’s CIFRE 2020/1206), Dreamquark, France-Travail, Institut Jean-Nicod, DEC-ENS Paris. 

Paul \'Egr\'e thanks Monash University (Philosophy Department) and the University of Melbourne (EEE Department) for their hospitality during the writing of this paper. We also thank three anonymous referees and  Xinghan Liu and Emiliano Lorini for their helpful comments. 

We also thank Branden Fitelson, Marc Fleurbaey, Maxilian Kiener, Ali Ozkes, Milo Phillips-Brown, Chris Russell, and the members of the HYBRINFOX and TRUSTEDNEWS consortia for valuable discussions. Special thanks to Johnathan Nguyen for his help with the review and publication of the package on GitHub.

\section*{Author Contributions}

The creation of the \fairdream\ package originates in the AI competition Veritas on AI fairness for financial services, organized by the Monetary Authority of Singapore in 2021, where \fairdream\ was selected as a finalist.  The package \fairdream\ was coded by Thomas Souverain, who also conducted the benchmark experiments reported in this paper, and created the repository of the results. The paper was conceptualized and written by Thomas Souverain and Paul \'Egr\'e, who share equal responsibility in its elaboration and successive revisions

\end{document}